%% file: main.tex
\documentclass[12pt,authoryear]{elsarticle}

\usepackage{graphicx}
\usepackage{amssymb}
\usepackage{lineno}
\usepackage{rotating}
\usepackage{fancyhdr}

\usepackage{color}
\definecolor{myred}{rgb}{.8,.0,.0}

\definecolor{myblue}{rgb}{.0,.0,.0}
\newcommand{\added}[1]{\textcolor{myblue}{#1}}

\journal{Current Opinion in Biomedical Engineering (accept), 10.1016/j.cobme.2018.12.005     }

\date{}

\begin{document}
\begin{frontmatter}

\title{Cats or CAT scans: transfer learning from natural or medical image source datasets?}


\author{Veronika Cheplygina}

\address{Eindhoven University of Technology}

\begin{abstract}
Transfer learning is a widely used strategy in medical image analysis. Instead of only training a network with a limited amount of data from the target task of interest, we can first train the network with other, potentially larger source datasets, creating a more robust model. The source datasets do not have to be related to the target task. For a classification task in lung CT images, we could use both head CT images, or images of cats, as the source. While head CT images appear more similar to lung CT images, the number and diversity of cat images might lead to a better model overall. In this survey we review a number of papers that have performed similar comparisons. Although the answer to which strategy is best seems to be ``it depends'', we discuss a number of research directions we need to take as a community, to gain more understanding of this topic.
\end{abstract}

\end{frontmatter}

\lfoot{Accepted to Current Opinion in Biomedical Engineering, DOI 10.1016/j.cobme.2018.12.005}



\section{Introduction}
\input{introduction.tex}

\section{Comparisons of source datasets} \label{sec:comparisons}
\input{papers_medical.tex}

\subsection{Public source datasets} \label{sec:datasets}
\input{datasets.tex}

\section{Discussion} \label{sec:discussion}
\input{discussion.tex}

\section*{Conflict of Interest}
The authors declare no conflict of interest.

\bibliographystyle{abbrvnat}
\bibliography{refs_main.bib}

\end{document}

%% file: introduction.tex
\begin{figure}
    \centering
    \includegraphics[width=\textwidth]{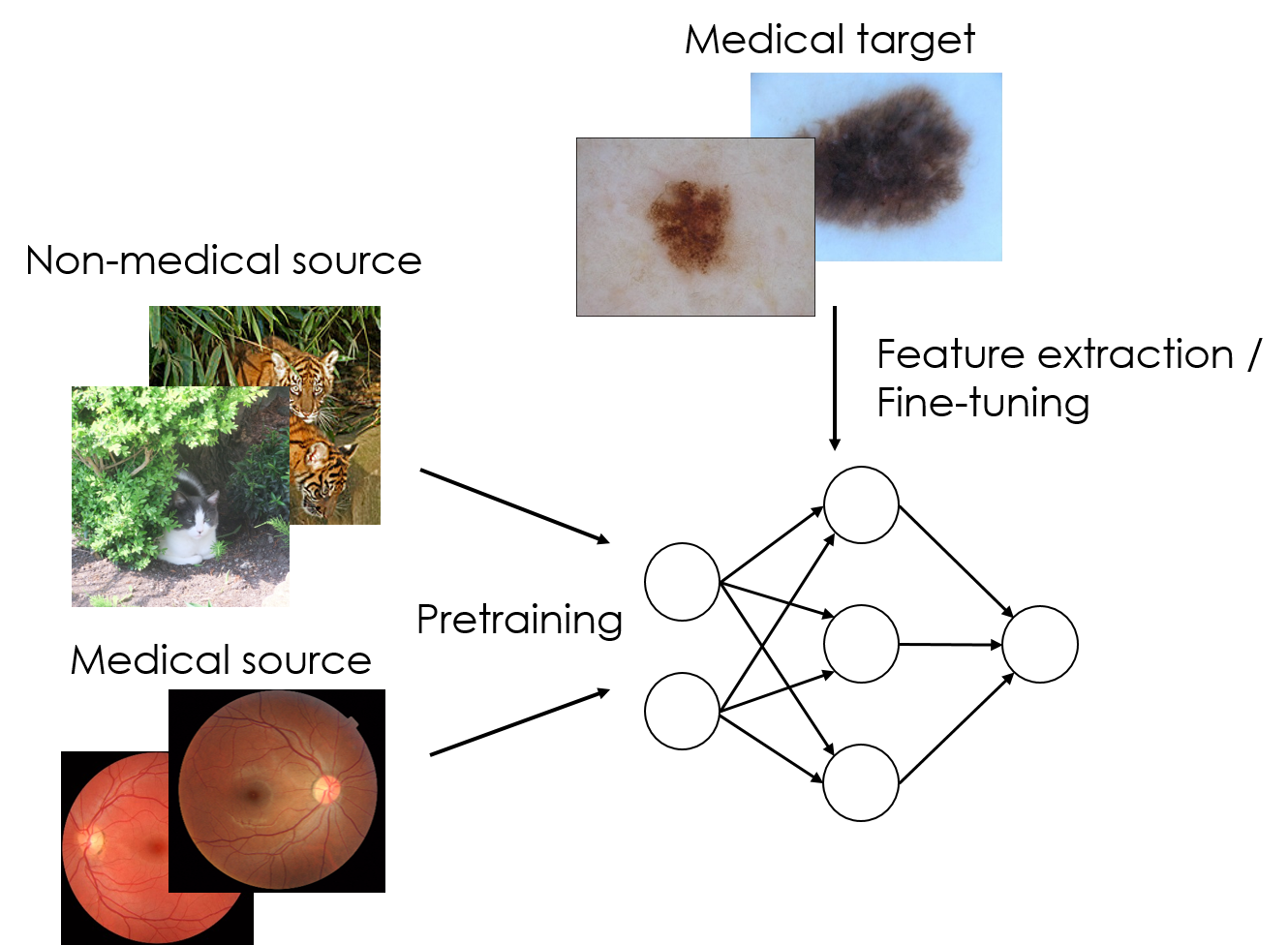}
    \caption{Transfer learning from non-medical or medical image datasets. A network is first trained on a source dataset. This network can then be used to for feature extraction or further training on the medical target data.}
    \label{fig:overview}
\end{figure}

In recent years transfer learning has become a popular technique for training machine learning classifiers~\citep{greenspan2016guest,litjens2017survey,cheplygina2018supervised}. The idea is to transfer information from one classification problem (the source) to the next (the target), thereby increasing the amount of data seen by the classifier. This is important for medical imaging, where datasets can be relatively small. In this review we look specifically at a type of transfer learning - training a network on one type of data, and then further training it on (a small amount of) possibly unrelated type of data. An illustration of this procedure is shown in Fig.~\ref{fig:overview}.

Training a neural network for a target dataset is typically achieved by one of three main strategies:
\begin{itemize}
\item Training the network ``from scratch'' or ``full training'', i.e. randomly initializing the weights and only using data from the target domain for training. In this case, no transfer learning is done.
\item Using ``off-the-shelf'' features, i.e. training a network on source data, using this pretrained network to extract features from the target data, and training another classifier, for example a support vector machine (SVM), on the extracted features. This is a type of transfer learning. 
\item Training with ``fine-tuning'', i.e. training a network on source data, then using this network to initialize the weights of a network that is further trained with target data. During training, some layers can be ``frozen'' so that their weights do not change. This is another type of transfer learning.
\end{itemize}
More details on each strategy can be found in \citep{litjens2017survey,yamashita2018convolutional}.

In transfer learning, the source problem may be seemingly unrelated to the target problem that is being solved. For example, ImageNet~\citep{russakovsky2015imagenet}, a large-scale dataset for object recognition, has been successfully used as source data for many medical imaging target tasks, with \citep{schlegl2014unsupervised,bar2015chest,ciompi2015automatic} among the earliest examples. Using other medical datasets as source data is less frequent, possibly because pretrained models are not as conveniently available as models trained on ImageNet, which are included in various toolboxes. It is therefore unclear whether pretraining on ImageNet is indeed the best strategy to choose for transfer learning in medical imaging. 

In this paper we review a number of papers which have used multiple source and/or target datasets, where the target datasets are from the medical imaging domain. The papers were selected by searching Google Scholar for ``transfer learning'' and ``medical imaging'' or ``biomedical imaging''. As of September 2018, this yielded close to 2K hits, over 90\% of which were from the last five years. We screened the results for papers where the title or abstract suggested that multiple source and/or target datasets have been used. Additionally, we screened the references of recent surveys~\citep{litjens2017survey,cheplygina2018supervised}, and screened the references and citations of the papers cited within this review. All selected papers were published between 2014 to 2018. A key change that happened in 2014, is being able to transfer from non-medical datasets, which often yielded good results and gave this area of research a boost.

Our goal is to get insights into what type of considerations should be made when choosing a source dataset for transfer learning. We first review the papers that compare different source data (Section~\ref{sec:comparisons}) and provide a summary of publicly available source datasets (Section~\ref{sec:datasets}). We then discuss several gaps in current literature and opportunities for future research in Section~\ref{sec:discussion}. 

%% file: papers_medical.tex
In this section we discuss the papers which provide insights on using non-medical or medical datasets for transfer learning. The papers are sorted by year, and then alphabetically.

\cite{schlegl2014unsupervised} is the earliest reference we are aware of doing transfer learning from non-medical data. \added{The application is} five-class classification of abnormalities in 2D slices of chest CT images. They pretrain an unsupervised convolutional restricted Boltzmann machine on different source datasets with 20K patches, and fine-tune an entire convolutional neural network (CNN) with varying sizes of lung patches. The target data is from 380 chest CT scans of the LTRC dataset~\citep{bartholomai2006lung}. The source data includes chest CT scans from LTRC, chest CT scans from a private dataset, brain CT scans from a private dataset, and natural images from the STL-10 dataset~\citep{coates2011analysis}, a subset of ImageNet. Natural images performed comparably or even slightly better than using only lung images. Brain images were less effective, possibly due to large homogeneous areas present in the scans, which are not present in more texture-rich lung scans. 

\cite{tajbakhsh2016convolutional} address four different applications: polyp detection in colonoscopy, image quality assessment in colonoscopy, pulmonary embolism detection in CT, and intima-media boundary segmentation in ultrasonography. \added{The authors} investigate full training and fine-tuning in a layerwise manner with AlexNet pretrained on ImageNet. 
Overall they observe that fine-tuning only the last layers performed worse than full training, but fine-tuning more layers was comparable to, or outperformed full training. 
Fine-tuning more layers was especially important for polyp detection and intima-media boundary segmentation, which the authors hypothesize are less similar to ImageNet than the other applications they examined. 

\cite{shin2016deep} address two tasks: thoraco-abdominal lymph node detection and interstitial lung disease (ILD) classification. CIFAR-10~\citep{krizhevsky2009learning} and ImageNet are used as source data. \added{Three strategies are compared:} training from scratch, off-the-shelf and fine-tuning strategies for different networks: CifarNet (trained on CIFAR-10), AlexNet (trained on ImageNet) and GoogLeNet. CifarNet is used only with the off-the-shelf strategy, AlexNet with all three, and GoogLeNet only with from-scratch and fine-tuning. 
For lymph node detection, the off-the-shelf strategy gives the worst results, but CifarNet outperforms AlexNet. Full training and fine-tuning lead to the best results, with fine-tuning being most beneficial for GoogLeNet. For ILD classification, AlexNet achieves similar performance with all three strategies, and for GoogLeNet fine-tuning is the most beneficial.
 
\cite{zhang2017automatic} address detection and classification of colorectal polyps in endoscopy images. They pretrain an eight-layer CNN and use the lower layers to extract features from the target data, which are then classified with an SVM. ImageNet and Places~\citep{zhou2017places} \added{are used as} as source datasets. Target datasets \added{include} a private endoscopy dataset with 2K images in three classes, and a public endoscopy video dataset~\citep{mesejo2016computer} from which 332 images in three classes \added{are extracted}. They hypothesize that Places has higher similarity between classes than ImageNet, which would help distinguish small differences in polyps. This indeed leads to higher recognition rates, also while varying other parameters of the classifier. 

\cite{cha2017bladder} predict the response to cancer treatment in the bladder of 82 patients using a five-layer CNN. They compare networks without transfer learning to two other source datasets: 60K natural images from CIFAR-10, and 160K bladder regions of interest (ROIs) from 81 patients from a previous study. \added{The experiments show} no statistically significant differences in the AUC values of two-fold cross-validation using these strategies. 

\cite{christodoulidis2017multisource} address classification of interstitial lung disease in patches of CT images. Six public texture source datasets \added{are used for} training a seven-layer network on each dataset and combining the networks in an ensemble. Individually, the source datasets result in networks with comparable performance, but the performance varies a lot depending on the number of layers transferred. The ensemble outperforms the individual networks. The ensemble also outperforms a network trained on the union of the datasets. 

\cite{menegola2017knowledge} address melanoma classification in skin lesion images. \added{Source data consists of} ImageNet and Kaggle diabetic retinopathy (DR)~\citep{graham2015kaggle}. The authors compare off-the-shelf, full training and fine-tuning strategies for a VGG network. They also investigate ``double transfer'': fine-tuning the pretrained ImageNet model on KaggleDR and only then on the target task. Fine-tuning outperforms off-the-shelf features when transferring from both sources. When transferring from ImageNet, off-the-shelf features outperform full training, but when transferring from KaggleDR, off-the-shelf features perform comparably with full training. Double transfer performs worse than transfer from ImageNet alone. This is in contrast to the hypothesis of the authors, that KaggleDR will lead to best results because of the visual similarity of the data.

\cite{ribeiro2017exploring} investigate pretraining and fine-tuning with nine different source datasets (natural images, texture images and endoscopy images) for classification of polyps in endoscopy images. Different from most other papers, they extract datasets of the same number of classes and images from the available types of data for the pretraining. \added{The experiments show} that texture datasets perform best as source data, but if the size of the source dataset is small, it is better to select a larger unrelated source dataset.

\cite{shi2018learning} address prediction of ocult invasive disease in ductal carcinoma in situ (DCIS) in mammography images of 140 patients. Three public datasets \added{are used} as the source data: ImageNet, texture dataset DTD~\citep{cimpoi2014describing} and dataset of mammography images INbreast~\citep{moreira2012inbreast}. \added{The authors} pretrain a 16-layer VGG network, extract off-the-shelf features from the target data using different network layers, and train a logistic regression classifier. They hypothesize that INbreast is most similar to the target data and will lead to the best results (and conversely, the least similar ImageNet will lead to the worst results), and report that the average AUCs are consistent with this hypothesis. 

\cite{du2018performance} address classification of 15K epithelium and stroma ROIs in 158 digital pathology images. ImageNet and Places are used as the source data. They extract off-the-shelf features from different layers of several architectures, where only AlexNet is trained on both sources. Comparing the AUCs of the AlexNet trained on ImageNet and Places, the layer used to extract the features (lower layers are better) has more influence than which data is used for pretraining.   

\cite{mormont2018comparison} focus on tissue classification. They argue that experiments are often carried out on a single dataset, therefore as target data eight tissue classification datasets with 1K to 30K images and two to ten classes \added{are used}. Seven architectures which are all trained on ImageNet \added{are compared}. \added{The method consists of} extracting features off-the-shelf or after fine-tuning, and training a supervised classifier. \added{The results} show that fine-tuning usually outperforms the other methods for any network, especially for multi-class datasets. The last layer is never the best for feature extraction, possibly because the features are too specific for natural images. Furthermore, the results do not suggest that larger datasets necessarily lead to better results - the smallest and the largest datasets lead to the best performances equally often.

\cite{lei2018deeply} address HEp-2 cell classification in the ICPR 2016 challenge as the target task~\citep{lovell2016international}. \added{Compared models include} a ResNet pretrained on ImageNet, and a ResNet pretrained on data from the earlier edition of the challenge, ICPR 2012~\citep{foggia2013benchmarking}. \added{The authors} hypothesize that pretraining on ICPR 2012 will lead to similar feature representations both in the lower and higher layers, and show that the network pretrained on ICPR 2012 data outperforms the ImageNet network.

\cite{wong2018building} focus on two tasks: three-class classification of brain tumors in 3D MR images and nine-class classification in 2D cardiac CTA images. They argue that pretrained ImageNet models are not suitable for medical target tasks because of unnecessary resizing of images, too large number of classes, and the absence of 3D information. \added{The classifier is} a modified U-Net which is first trained on a segmentation task on the same data, using either manual segmentations or segmentations generated with a simple thresholding method. In tumor classification, where ImageNet is not tested due to the 3D nature of the images, pretraining both with manual and thresholded segmentations outperforms training a network from scratch. In cardiac image classification, pretraining with manual segmentations gives the best results. Pretraining on ImageNet outperforms pretraining on thresholded segmentations. Pretraining on ImageNet also outperforms training from scratch, but only for low training sizes.

%% file: datasets.tex
A list of publicly available source datasets used in papers comparing multiple sources, but focusing on medical target tasks, is presented in Table~\ref{tab:datasets}. Imagenet is a popular choice, although some papers use other object recognition datasets such as CIFAR-10. Several papers use texture datasets, of which a variety is available. Only a few medical source datasets are listed, often because a private medical dataset is used.  


\begin{sidewaystable} 
\centering
\footnotesize
\begin{tabular}{l l l l p{6cm}}
Data & Type & Images & Classes & Used in \\

ImageNet \citep{russakovsky2015imagenet} & object recognition & 1.2M & 1K & \cite{tajbakhsh2016convolutional} \\
 & & & & \cite{shin2016deep}\\
 & & & & \cite{menegola2017knowledge}\\
 & & & & \cite{zhang2017automatic}\\
 & & & & \cite{du2018performance}\\
 & & & & \cite{mormont2018comparison} \\
 & & & & \cite{lei2018deeply} \\
 & & & & \cite{wong2018building} \\

STL-10 \cite{coates2011analysis} & object recognition & 100K & 10 & \cite{schlegl2014unsupervised} \\

Places \citep{zhou2017places} & scene recognition & 2.5M & 205 & \cite{zhang2017automatic}\\
& & & & \cite{du2018performance}\\
DTD \citep{cimpoi2014describing} & texture classification & 5.6K & 47 & \cite{ribeiro2017exploring} \\
& & & & \cite{shi2018learning}\\
& & & & \citep{christodoulidis2017multisource} \\
ALOT \citep{burghouts2009material} & texture classification & 28K & 250 & \cite{ribeiro2017exploring} \\
& & & & \citep{christodoulidis2017multisource} \\
KTH-TIPS \citep{dana1999reflectance} & texture classification & 810 & 10 & \cite{ribeiro2017exploring}\\
& & & & \citep{christodoulidis2017multisource} \\
CIFAR-10 \citep{krizhevsky2009learning} & object classification & 60K & 10 & \cite{shin2016deep}\\
& & & & \cite{cha2017bladder}\\

FMD \citep{sharan2009material} & texture classification & 1K & 10 & \cite{christodoulidis2017multisource} \\
KTB \citep{kylberg2011kylberg} & texture classification & 4.5K & 27 & \cite{christodoulidis2017multisource} \\
UIUC \citep{lazebnik2005sparse} & texture classification & 1K & 25 & \cite{christodoulidis2017multisource} \\

CALTECH-101 \citep{li2006one} & object recognition & 9K & 101 & \cite{ribeiro2017exploring}\\
COREL-1000 & scene recognition & 1K & 10 & \cite{ribeiro2017exploring} \\ 

Kaggle-DR \cite{graham2015kaggle} & retinal image classification & 35K & 2 & \cite{menegola2017knowledge}\\
INbreast \citep{moreira2012inbreast} & breast lesion classification & 410 & 2 & \cite{shi2018learning}\\
ICPR 2012 \citep{foggia2013benchmarking} & HEp-2 cell classification & 1.5K & 6 & \cite{lei2018deeply} \\

\end{tabular}
\caption{Overview of public datasets used as source data. In several cases, the sample size is larger in cases when networks are trained on patches extracted from images, rather than entire images.}
\label{tab:datasets}
\end{sidewaystable}

\normalsize

%% file: discussion.tex
We have summarized several papers which use medical or non-medical data as source data and apply the classifier on medical target data. A limitation is that such papers are difficult to discover - other than ``transfer learning'', which returns over 2K results when combined with ``medical imaging'' on Google Scholar, we have not been able to find keywords that identify when different source of datasets have been used. We encourage readers to notify us of any papers that were not included, but also investigate this phenomenon.

\subsection{Summary of results}
The results of the comparisons point in different directions. We now group the papers into ``non-medical is best'', ``medical wins'', ``no difference'' and ``inconclusive''. 

Out of twelve papers in Section~\ref{sec:comparisons}, three papers conclude that natural images outperform medical images. \cite{schlegl2014unsupervised,menegola2017knowledge} find natural images (STL-10 and ImageNet respectively) more effective than medical. \cite{ribeiro2017exploring} have most success with texture images, compared to natural or medical images. We consider these texture images as non-medical.  

Five papers can be seen as voting for medical images. Three papers~\citep{shi2018learning,lei2018deeply,wong2018building} have clear conclusions  that better results are achieved with medical source data. Next to this, there are two ``minor'' votes for medical images. \cite{tajbakhsh2016convolutional} only use ImageNet as source data, but they use different target datasets, and conclude that ImageNet is worse for datasets that are less similar to medical images. This could be seen as a vote for medical datasets, since these would be most similar to the target data. A similar intuition holds for \cite{zhang2017automatic}, who use ImageNet and Places as the source data, and find that Places is better because of its higher similarity to medical data.

In two papers, there are no clear differences between source datasets. \cite{cha2017bladder} find no significant differences between using different (medical and non-medical) sources. \cite{du2018performance} do not demonstrate consistent differences between ImageNet and Places as source data. 

From three of the papers, we cannot reason whether non-medical or medical is better because of the source choices, but they do provide other relevant insights. \cite{shin2016deep} use two natural datasets as sources and show that a larger natural image dataset is not necessarily better. The same holds for \cite{christodoulidis2017multisource}, who use only texture datasets as sources, but show that training on the union of the datasets does not improve the results. Similarly, \cite{mormont2018comparison} show that both smaller and larger medical sources can be successful as source data.

\subsection{Limitations}
It is difficult to compare these results directly because of differences in how transfer learning is implemented. Examples of variation include the subset of the source data that is used, the architecture of the network, and how the transfer was implemented, both in terms of strategy (off-the-shelf or fine-tuning) and which layers were used for the transfer. Furthermore, since we could only find twelve relevant comparisons, a difference of several votes for one or the other approach could be due to chance.

Another issue is that the target datasets in medical imaging can be very small, and it is not clear if the results would generalize to another similar dataset. Methods are sometimes compared by looking only at a single run of each method, or at an average over multiple runs, but without considering possible variability in such performances. A recent paper comparing medical image challenges~\cite{maierhein2018winner} shows that in such conditions, rankings of algorithms can easily change, for example if a slightly different metric is used. Most papers we surveyed performed no statistical significance tests - if this was the case, perhaps the conclusions would be different.

\subsection{Opportunities for further research}
There are opportunities in doing more systematic comparisons. One direction is to use more of the available datasets, both from the non-medical and medical domains. It would be informative to vary the number of images and number of classes in the data, similar to \citep{ribeiro2017exploring}. Also of interest would be comparing different tasks, such as segmentation and classification, involving the same images, similar to \citep{wong2018building}. 

The number of public medical source datasets is rather low. 
A strategy that could be helpful to counteract this, but seems underexplored, is unsupervised pretraining. This would allow the use of larger unlabeled medical datasets, which may be only weakly labeled. Many such datasets are already publicly available, for example on grand-challenge.org. Another way to increase the number of source datasets would be to share pretrained models, which would also allow transfer learning from private datasets, without sharing the data itself. The feasibility of this approach with respect to privacy and intellectual property of the data would need to be investigated first. 

Similarity of datasets is often used to hypothesize about which source data will be best, but definitions of similarity differ.  For example, \cite{menegola2017knowledge} discuss similarity in terms of visual similarities of the images, \cite{lei2018deeply} discuss similarity in terms of feature representations. In computer vision, other definitions may be used - for example, \cite{azizpour2015generic} investigate transfer from ImageNet and Places to 15 other datasets, and define similarity in terms of the number and variety of the classes. Given a definition of similarity, it remains a question which datasets would be best to use for pretraining. Arguably, the most similar dataset to the target dataset, is the target dataset itself, which might not add any additional information. Investigating how to represent datasets in a feature space (one example can be found in ~\cite{cheplygina2017exploring}) or how to directly define dataset similarity is an important point of investigation.

Instead of considering only the similarity of the source data, perhaps the diversity of the source data is also an important factor. Instead of selecting one dataset as the source, it might be a good strategy to use an ensemble, similar to ~\citep{christodoulidis2017multisource}. It is in fact surprising, given that top performing methods in challenges are often ensembles, that this strategy was not investigated in the papers we reviewed. It might be the case that it isn't better to use non-medical or medical data - the answer to the question posed by the title, might simply be ``both''. 

\subsection{Concluding remarks} 

In conclusion, we looked at twelve papers which compare various source and/or target datasets from different domains. A similar number of papers found that non-medical or medical data was better, with a slight advantage for the medical data. Several papers showed that larger datasets are not necessarily better. Since datasets, dataset sizes, architectures and transfer strategies vary between comparisons, and results are often based on a few datasets without significance testing, we urge the community to conduct larger systematic comparisons into this important topic.